\documentclass[table]{article}
\pdfoutput=1
\usepackage{iclr2017_conference,times}

\usepackage[utf8]{inputenc} 
\usepackage[T1]{fontenc}    
\usepackage{hyperref}       
\usepackage{url}            
\usepackage{booktabs}       
\usepackage{amsfonts}       
\usepackage{nicefrac}       
\usepackage{microtype}      

\newcolumntype{C}[1]{>{\centering\let\newline\\\arraybackslash\hspace{0pt}}m{#1}}

\usepackage{amsmath}
\usepackage{graphicx}
\usepackage{multirow}  
\usepackage{bm}
\usepackage{verbatim}
\usepackage{subcaption}
\usepackage{bmpsize}
\usepackage{subcaption,siunitx,booktabs}

\title{Query-Reduction Networks\\
for Question Answering}

\author{
  Minjoon Seo$^1$\qquad Sewon Min$^2$\qquad Ali Farhadi$^{1,3}$\qquad Hannaneh Hajishirzi$^1$ \\
  University of Washington$^1$, Seoul National University$^2$, Allen Institute for Artificial Intelligence$^3$ \\
  \texttt{\{minjoon, ali, hannaneh\}@cs.washington.edu, shmsw25@snu.ac.kr}
}

\iclrfinalcopy

\begin{document}

\maketitle

\begin{abstract}

In this paper, we study the problem of question answering when reasoning over multiple facts is required. 
We propose Query-Reduction Network (QRN), a variant of Recurrent Neural Network (RNN) that effectively handles both short-term (local) and long-term (global) sequential dependencies to reason over multiple facts. 
QRN considers the context sentences as a sequence of state-changing triggers, and \emph{reduces} the
original query to a more informed query as it observes each trigger (context sentence) through time. 
Our experiments show that QRN produces the state-of-the-art results in  bAbI QA and dialog tasks, and in a real goal-oriented dialog dataset. In addition, QRN formulation allows parallelization on RNN's time axis, saving an order of magnitude in time complexity for training and inference.  

\end{abstract}

\section{Introduction}\label{sec:intro}

In this paper, we address the problem of  question answering (QA) when reasoning over multiple facts is required. 
For example, consider we know that \texttt{Frogs eat insects} and \texttt{Flies are insects}. 
Then answering \texttt{Do frogs eat flies?} requires reasoning over both of the above facts. Question answering, more specifically context-based QA, has been extensively studied in machine comprehension tasks~\citep{MCTest,hermann2015teaching,hill2015goldilocks,squad}. However, most of the datasets are primarily focused on lexical and syntactic understanding, and hardly concentrate on inference over multiple facts. 
Recently, several datasets aimed for testing multi-hop reasoning have emerged; among them are story-based QA~\citep{babi} and the dialog task~\citep{bordes2016learning}.

Recurrent Neural Network (RNN) and its variants, such as Long Short-Term Memory (LSTM)~\citep{lstm} and Gated Recurrent Unit (GRU)~\citep{GRU}, are popular choices for modeling natural language. However, when used for multi-hop reasoning in question answering, purely RNN-based models have shown to perform poorly~\citep{babi}. This is largely due to the fact that RNN's internal memory is inherently unstable over a long term. For this reason, most recent approaches in the literature have mainly relied on global attention mechanism and shared external memory~\citep{memN2N,NR,DMN+,graves2016hybrid}.
The attention mechanism allows these models to focus on a single sentence in each layer.
They can sequentially read multiple relevant sentences from the memory with multiple layers to perform multi-hop reasoning. However, one major drawback of these standard attention mechanisms is that they are insensitive to the time step (memory address) of the sentences when accessing them.

\begin{figure}[t]
\centering
\begin{subfigure}[htbp]{0.242\textwidth}
\includegraphics[width=\textwidth]{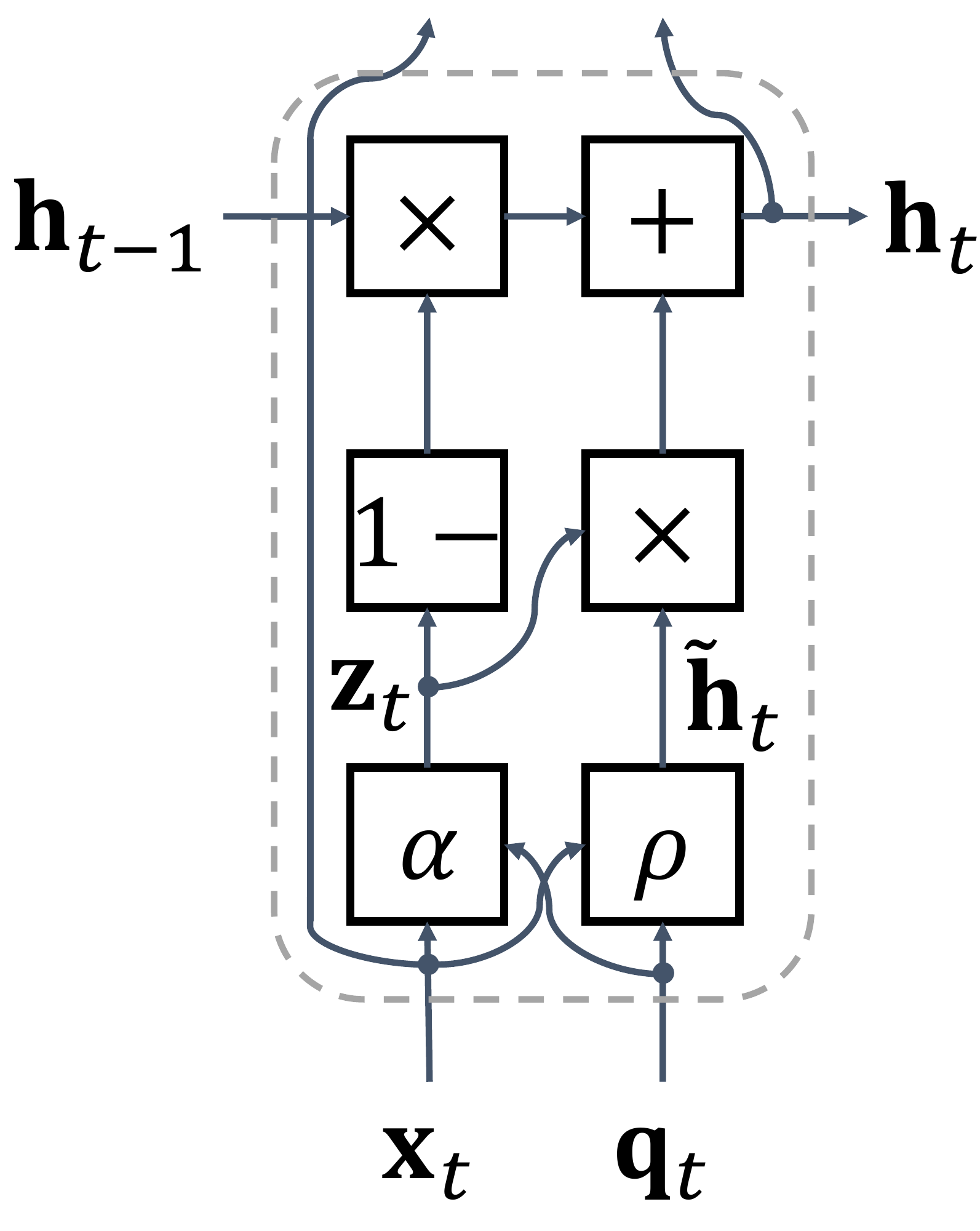}
\caption{  QRN unit }
\label{fig:unit}
\end{subfigure}
\begin{subfigure}[htbp]{0.565\textwidth}
\centering
\includegraphics[width=\textwidth]{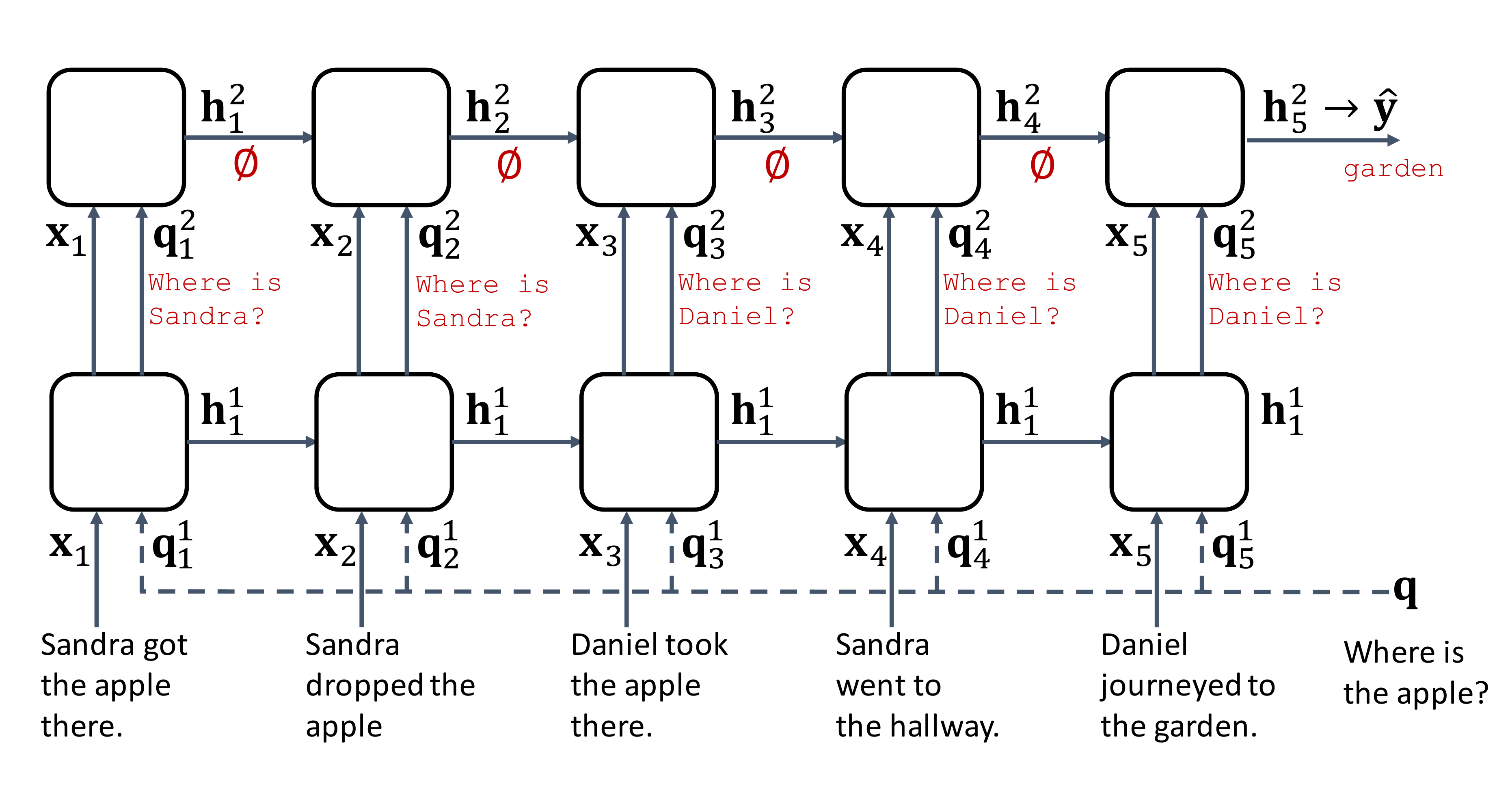}
\caption{2-layer QRN}
\label{fig:qrn_ex}
\end{subfigure}
\begin{subfigure}[htbp]{0.143\textwidth}
\centering
\includegraphics[width=\textwidth]{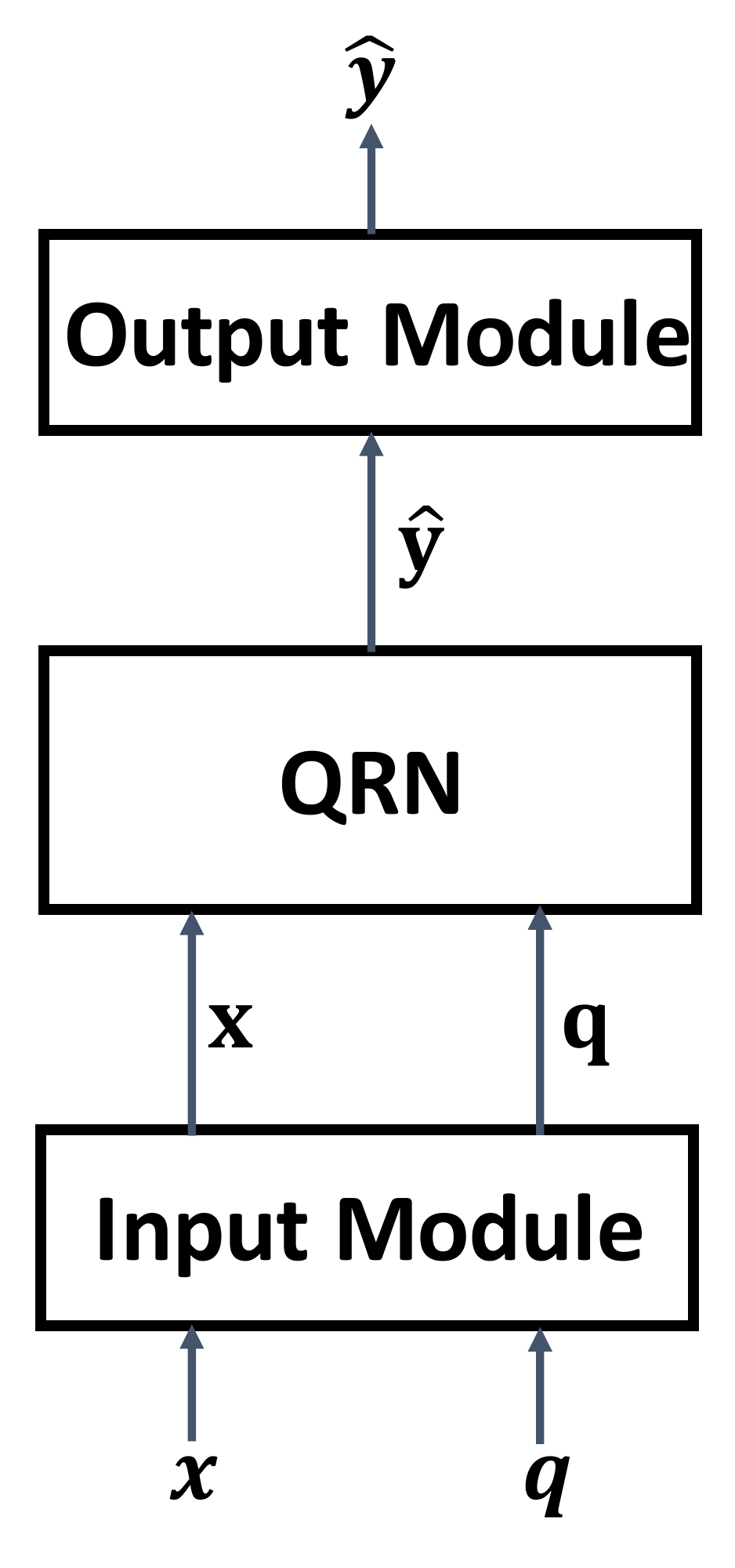}
\caption{Overview}
\label{fig:overview}
\end{subfigure}
\caption{\small (\ref{fig:unit}) QRN unit, (\ref{fig:qrn_ex})  2-layer QRN on 5-sentence story, and (\ref{fig:overview}) entire QA system (QRN and input / output modules). 
    ${\bm x}, {\bm q}, \hat{\bm y}$ are the story, question and predicted answer in natural language, respectively.
    ${\bf x}=\langle {\bf x}_1, \ldots , {\bf x}_T \rangle, {\bf q}, \hat{\bf y}$ are their corresponding vector representations (upright font).
    $\alpha$ and $\rho$ are update gate and reduce functions, respectively.
    ${\hat {\bf y}}$ is  assigned to be ${\bf h}^2_5$, the local query at the last time step in the last layer.
    Also, red-colored text is the inferred meanings of the vectors (see `Interpretations' of Section~\ref{subsec:results}). }
\label{fig:model}
\end{figure}

Our proposed model, Query-Reduction Network\footnote{Code is publicly available at: \url{seominjoon.github.io/qrn/}}(QRN), is a single recurrent unit that addresses the long-term dependency problem of most RNN-based models by simplifying the recurrent update, while taking the advantage of RNN's capability to model sequential data (Figure~\ref{fig:model}).
QRN considers the context sentences as a sequence of state-changing triggers, and transforms (\emph{reduces}) the original query to a more informed query as it observes each trigger through time. 
For instance in Figure~\ref{fig:qrn_ex}, 
the original question, \texttt{Where is the apple?}, cannot be directly answered by any single sentence from the story. After observing the first sentence, \texttt{Sandra got the apple there}, QRN transforms the original question to a reduced query \texttt{Where is Sandra?}, which is presumably easier to answer than the original question given the context provided by the first sentence.\footnote{This mechanism is akin to logic regression in situation calculus~\citep{reiter}.}
Unlike RNN-based models, QRN's candidate state ($\tilde{\bf h}_t$ in Figure~\ref{fig:unit}) does not depend on the previous hidden state (${\bf h}_{t-1}$). 
Compared to memory-based approaches~\citep{memNet,memN2N,NR,DMN,DMN+}, QRN can better encodes \emph{locality} information because  it does not use a global memory access controller (circle nodes in Figure~\ref{fig:models}), and the query updates are performed locally.

In short, the main contribution of QRN is threefold. 
First, QRN is a simple variant of RNN that {\it reduces} the query given the context sentences in a differentiable manner.
Second, QRN is situated between the attention mechanism and RNN, effectively handling time dependency and long-term dependency problems of each technique, respectively. 
Hence it is well-suited for sequential data with both local and global interactions (note that QRN is \emph{not} the replacement of RNN, which is arguably better for modeling complex local interactions).
Third, unlike most RNN-based models, QRN can be parallelized over time by computing candidate reduced queries ($\tilde{\bf h}_t$) directly from local input queries (${\bf q}_t$) and context sentence vectors (${\bf x}_t$).
In fact, the parallelizability of QRN implies that QRN does not suffer from the vanishing gradient problem of RNN, hence effectively addressing the long-term dependency.
We experimentally demonstrate these contributions by achieving the state-of-the-art results on story-based QA and interactive dialog datasets.

\section{Model}\label{sec:qrn}

In story-based QA (or dialog dataset), the input is the \emph{context} as a sequence of sentences (story or past conversations) and a \emph{question} in natural language (equivalent to the user's last utterance in the dialog). 
The output is the predicted answer to the question in natural language (the system's next utterance in the dialog).
The only supervision provided during training is the answer to the question. 



In this paper we particularly focus on end-to-end solutions, i.e.,  the only supervision comes from questions and answers, and we restrain from using manually defined rules or external language resources, such as lexicon or dependency parser. 
Let $\langle \bm{x}_1, \ldots, \bm{x}_T\rangle$ denote the sequence of sentences,  where $T$ is the number of sentences in the story, and let $\bm{q}$ denote the question. 
Let $\hat{\bm{y}}$ denote the predicted answer, and ${\bm{y}}$ denote the true answer.
Our proposed system for end-to-end QA task is divided into three modules (Figure~\ref{fig:overview}): input module, QRN layers, and output module.

\paragraph{Input module.} 
Input module maps each sentence $\bm{x}_t$ and the question $\bm{q}$ to $d$-dimensional vector space, ${\bf x}_t \in \mathbb{R}^d$ and ${\bf q}_t \in \mathbb{R}^d$.
We adopt a previous solution for the input module (details in Section~\ref{sec:qa}).

\paragraph{QRN layers.} 
QRN layers use the sentence vectors and the question vector from the input module to obtain the predicted answer in vector space, $\hat{\bf y} \in \mathbb{R}^d$. 
A QRN layer refers to the recurrent application of a QRN unit, which can be considered as a variant of RNN with two inputs, two outputs, and a  hidden state (reduced query), all of which operate in vector space.
The details of the QRN module is explained throughout this section (\ref{sec:model}, \ref{sec:var}).

\paragraph{Output module.}
Output module maps $\hat{\bf y}$ obtained from QRN to a natural language answer $\hat{\bm y}$. 
Similar to the input module, we adopt a standard solution for the output module (details in Section~\ref{sec:qa}).

We first formally define the base model of a QRN unit, and then we explain how we connect the input and output modules to it (Section~\ref{sec:model}).
We also present a few extensions to the network that can improve QRN's performance (Section~\ref{sec:var}).
Finally, we show that QRN can be parallelized over time, giving computational advantage over most RNN-based models by one order of magnitude (Section~\ref{sec:par}).

\subsection{QRN Unit}\label{sec:model}
As an RNN-based model, QRN is a single recurrent unit that updates its hidden state (reduced query) through time and layers. Figure~\ref{fig:unit} depicts the schematic structure of a QRN unit, and Figure~\ref{fig:qrn_ex} demonstrates how layers are stacked.
A QRN unit accepts two inputs (\emph{local} query vector ${\bf q}_t \in \mathbb{R}^d$ and sentence vector ${\bf x}_t \in \mathbb{R}^d$), and two outputs (reduced query vector ${\bf h}_t \in \mathbb{R}^d$, which is similar to the hidden state in RNN, and the sentence vector ${\bf x}_t$ from the input without modification).
The local query vector is not necessarily identical to the original query (question) vector ${\bf q}$.
In order to compute the outputs, we use \emph{update gate} function $\alpha: \mathbb{R}^d \times \mathbb{R}^d \rightarrow [0, 1]$ and \emph{reduce} function $\bm{\rho}: \mathbb{R}^d \times \mathbb{R}^d \rightarrow \mathbb{R}^d$. 
Intuitively, the update gate function measures the relevance between the sentence and the local query and is used to update the hidden state. 
The reduce function transforms the local query input to a candidate state which is a new reduced (easier) query given the sentence.
The outputs are calculated with the following equations:
\begin{eqnarray}\label{eq:z}
&& z_t= \alpha({\bf x}_t, {\bf q}_t) = \sigma({\bf W}^{(z)} ({\bf x}_t \circ {\bf q}_t) + b^{(z)})\\
&&{\tilde {\bf h}}_t = {\bm \rho}({\bf x}_t, {\bf q}_t) = \tanh({\bf W}^{({\bf h})}[{\bf x}_t; {\bf q}_t] + {\bf b}^{({\bf h})}) \label{eq:ht}\\
&&{\bf h}_t = z_t {\tilde {\bf h}}_t + (1 - z_t) {\bf h}_{t-1}\label{eq:h}
\end{eqnarray}
where $z_t$ is the scalar update gate, ${\tilde {\bf h}}_t$ is the candidate reduced query, and ${\bf h}_t$ is the final reduced query at time step $t$,
$\sigma(\cdot)$ is sigmoid activation, $\tanh(\cdot)$ is hyperboolic tangent activation (applied element-wise),
${\bf W}^{(z)} \in \mathbb{R}^{1 \times d}$, ${\bf W}^{({\bf h})} \in \mathbb{R}^{d \times 2d}$ are weight matrices, 
$b^{(z)} \in \mathbb{R}$, ${\bf b}^{({\bf h})} \in \mathbb{R}^d$ are bias terms,
$\circ$ is element-wise vector multiplication, and $[;]$ is vector concatenation along the row.
As a base case, ${\bf h}_0 = {\bf 0}$.
Here we have explicitly defined $\alpha$ and ${\bm \rho}$,
but they can be any reasonable differentiable functions.

The update gate is similar to the global attention mechanism~\citep{memN2N,DMN+} in that it measures the similarity between the sentence (a memory slot) and the query.
However, a significant difference is that the update gate is computed using sigmoid ($\sigma$) function on the current memory slot only (hence internally embedded within the unit), whereas the global attention is computed using $\mathrm{softmax}$ function over the entire memory (hence globally defined).
The update gate can be rather considered as \emph{local sigmoid} attention.

\paragraph{Stacking layers} We just showed the single-layer case of QRN, but QRN with multiple layers is able to perform reasoning over multiple facts more effectively, as shown in the example of Figure~\ref{fig:qrn_ex}.
In order to stack several layers of QRN, the outputs of the current layer are used as the inputs to the next layer. 
That is, using superscript $k$ to denote the current layer's index (assuming $1$-based indexing), we let 
${\bf q}^{k+1}_t = {\bf h}^k_t$.
Note that ${\bf x}_t$ is passed to the next layer without any modification, so we do not put a layer index on it.

\paragraph{Bi-direction.}
So far we have assumed that QRN only needs to look at past sentences, whereas
often times, query answers can depend on future sentences.
For instance, consider a sentence ``John dropped the football.'' at time $t$.
Then, even if there is no mention about the ``football'' in the past (at time $i < t$), 
it can be implied that ``John'' has the ``football'' at the current time $t$. 
In order to incorporate the future dependency, we obtain $\overrightarrow{{\bf h}}_t$ and $\overleftarrow{{\bf h}}_t$ in both forward and backward directions, respectively, using Equation~\ref{eq:h}. We then  add them together to get ${\bf q}_t$ for the next layer. That is,
\begin{equation}
{\bf q}^{k+1}_t = \overrightarrow{{\bf h}}^k_t + \overleftarrow{{\bf h}}^k_t
\end{equation}
for layer indices $1 \leq k \leq K-1$.
Note that  the variables ${\bf W}^{(z)}, b^{(z)}, {\bf W}^{({\bf h})}, {\bf b}^{({\bf h})}$ are shared between the two directions.

\paragraph{Connecting input and output modules.}
Figure~\ref{fig:overview} depicts how QRN is connected with the input and output modules. 
In the first layer of QRN, ${\bf q}^1_t = {\bf q}$ for all $t$, where ${\bf q}$ is obtained from the input module by processing the natural language question input ${\bm q}$.
${\bf x}_t$ is also obtained from ${\bm x}_t$ by the same input module.
The output at the last time step in the last layer is passed to the output module.
That is, ${\hat {\bf y}} = {\bf h}^K_t$ where $K$ represent the number of layers in the network.
Then the output module gives the predicted answer ${\hat {\bm y}}$ in natural language.

\subsection{Extensions}\label{sec:var}
Here we introduce a few extensions of QRN, and later in our experiments, we test QRN's performance with and without each of these extensions.

\paragraph{Reset gate.} Inspired by GRU~\citep{GRU}, we found that it is useful to allow the QRN unit to reset (nullify) the candidate reduced query (i.e., ${\tilde {\bf h}}_t$) when necessary.
For this we use a \emph{reset gate} function $\beta : \mathbb{R}^d \times \mathbb{R}^d \rightarrow [0,1]$, which can be defined similarly to the update gate function:
\begin{equation}\label{eq:r}
r_t = \beta({\bf x}_t, {\bf q}_t) = \sigma({\bf W}^{(r)} ({\bf x}_t \circ {\bf q}_t) + b^{(r)})
\end{equation}
where ${\bf W}^{(r)} \in \mathbb{R}^{1 \times d}$ is a weight matrix, and $b^{(r)} \in \mathbb{R}$ is a bias term. 
Equation~\ref{eq:h} is rewritten as
\begin{equation}\label{eq:hr}
{\bf h}_t = z_t r_t {\tilde {\bf h}}_t + (1 - z_t) {\bf h}_{t-1}.
\end{equation}
Note that we do not use the reset gate in the last layer.

\paragraph{Vector gates.} As in LSTM and GRU, update and reset gates can be vectors instead of scalar values for fine-controlled gating.
For vector gates, we modify the row dimension of weights and biases in Equation~\ref{eq:z} and~\ref{eq:r} from $1$ to $d$.
Then we obtain ${\bf z}_t, {\bf r}_t \in \mathbb{R}^d$ (instead of $z_t, r_t \in \mathbb{R})$, and these can be element-wise multiplied ($\circ$) instead of being broadcasted in Equation~\ref{eq:h} and~\ref{eq:hr}.


\section{Parallelization}\label{sec:par}
An important advantage of QRN is that the recurrent updates in Equation~\ref{eq:h} and~\ref{eq:r}  can be computed in parallel across time. 
This is in contrast with most RNN-based models that cannot be parallelized, where computing the candidate hidden state  at time $t$ explicitly requires the previous hidden state.
In QRN, the final reduced queries (${\bf h}_t$) can be decomposed into computing over candidate reduced queries (${\tilde {\bf h}}_t$), without looking at the previous reduced query.
Here we primarily show that the query update in Equation~\ref{eq:h} can be parallelized by rewriting the equation with matrix operations. 
The extension to Equation~\ref{eq:r} is straightforward.
The proof for QRN with vector gates is shown in Appendix~\ref{sec:app-b}.
The recursive definition of Equation~\ref{eq:h} can be explicitly written as
\begin{equation}\label{eq:exp}
    {\bf h}_t 
    = \sum_{i=1}^{t} \left[\prod_{j=i+1}^{t}  1-z_j \right] z_i \tilde{{\bf h}}_i
    = \sum_{i=1}^{t} \exp \left \{\sum_{j=i+1}^{t} \log \left(1 - z_j \right) \right \}
    z_i \tilde{{\bf h}}_i.
\end{equation}
Let $b_i = \log(1-z_i)$ for brevity.
Then we can rewrite Equation~\ref{eq:exp} as the following equation:

\begin{equation}\label{eq:mat}
\begin{pmatrix}
{\bf h}_1^\top \\
{\bf h}_2^\top \\
{\bf h}_3^\top \\
\vdots \\
{\bf h}_T^\top
\end{pmatrix}
 = \left[
 \exp\left\{
 \begin{pmatrix}
 0 & -\infty & -\infty & \ldots & -\infty \\
 b_2  & 0 & -\infty & \ldots & -\infty\\
 b_2 +b_3  & b_3 & 0 & \ldots & -\infty\\
 \vdots & \vdots & \vdots & \ddots & \vdots\\
 \sum_{j=2}^T b_j & \sum_{j=3}^T b_j & \sum_{j=4}^T b_j &\ldots & 0
 \end{pmatrix}\right\}
\right]
\begin{pmatrix}
z_1 \tilde{{\bf h}}^\top_1 \\
z_2\tilde{{\bf h}}^\top_2 \\
z_3\tilde{{\bf h}}^\top_3 \\
\vdots \\
z_T\tilde{{\bf h}}^\top_T
\end{pmatrix}
\end{equation}
Let ${\bf H} = [{\bf h}_1^\top ; \ldots ; {\bf h}_T^\top]$ 
be a $T$-by-$d$ matrix where the transposes ($\top$) of the column vectors ${\bf h}_t$ are concatenated across row.
We similarly define $\tilde{{\bf H}}$ from $\tilde{{\bf h}}_t$. 
Also, let ${\bf z}=[z_1; \ldots ; z_T]$ and ${\bf b}=[0; b_2; \dots ; b_T]$ be column vectors (note that we use $0$ instead of $b_1$).
Then Equation~\ref{eq:mat} is: 
\begin{equation}\label{eq:short}
{\bf H} = \left[ {\bf L} \circ \exp \left( {\bf L}\left[{\bf B} \circ {\bf L}'\right]\right) \right]
 \left[ {\bf Z} \circ \tilde{{\bf H}} \right]
\end{equation}
where ${\bf L}, {\bf L}' \in \mathbb{R}^{T \times T}$ are lower and \emph{strictly} lower triangular matrices of $1$'s, respectively, 
$\circ$ is element-wise multiplication, and
${\bf B}$ is a matrix where $T$ ${\bf b}$'s are tiled across the column, i.e. ${\bf B} = [{\bf b}, \dots , {\bf b}] \in \mathbb{R}^{T \times T}$, 
and similarly ${\bf Z} = [{\bf z}, \dots , {\bf z}] \in \mathbb{R}^{T \times d}$. 
All implicit operations are matrix multiplications.
With reasonable $N$ (batch size), $d$ and $T$ (e.g. $ N,d,T=100$), matrix operations in Equation~\ref{eq:short} can be comfortably computed in most modern GPUs.

\section{Related Work}\label{sec:rel}



\begin{figure}[t]

\begin{subfigure}[htbp]{0.32\textwidth}
\centering
\includegraphics[width=\textwidth]{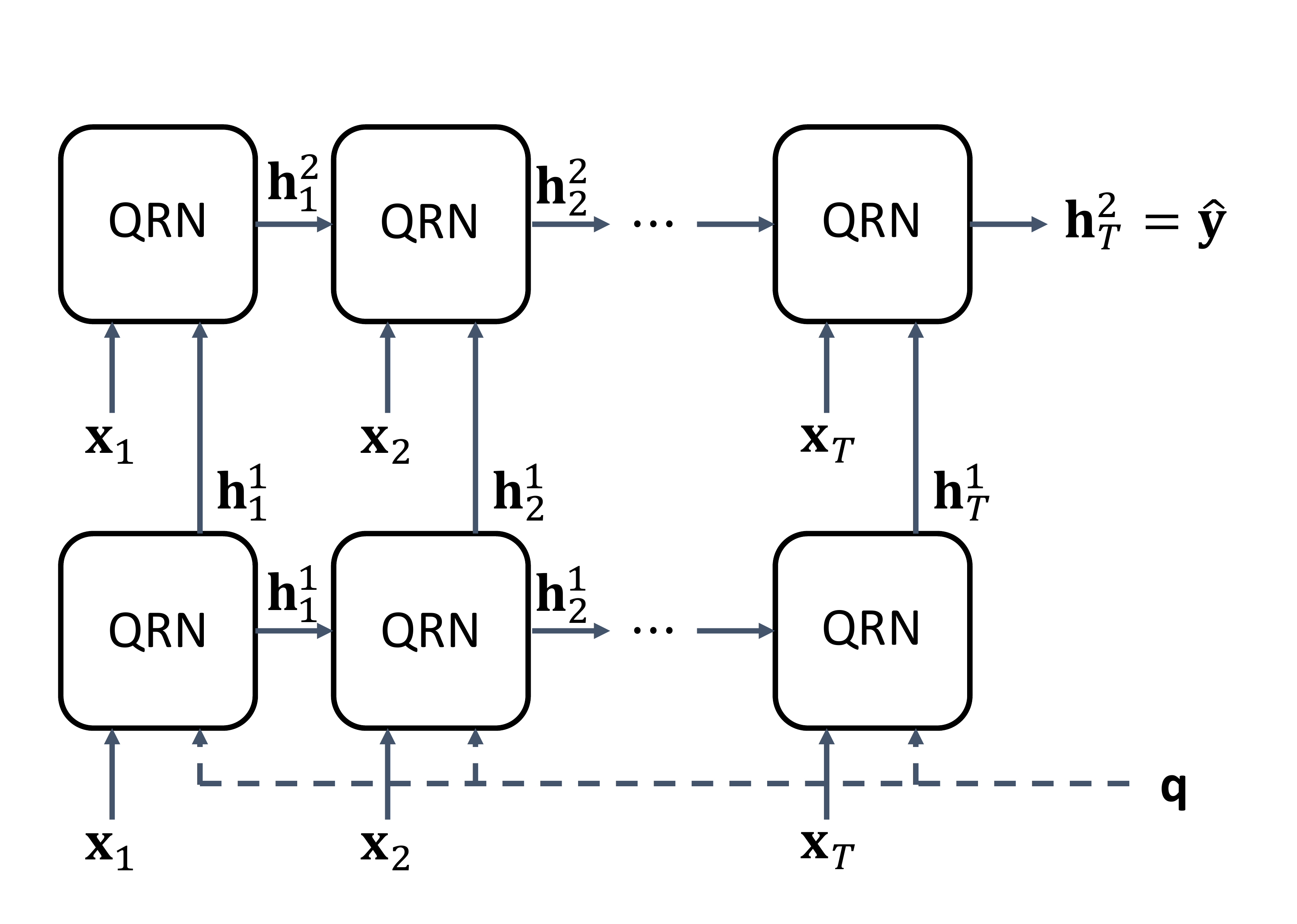}
\caption{  QRN }
\label{fig:qrn}
\end{subfigure}
\begin{subfigure}[htbp]{0.32\textwidth}
\centering
\includegraphics[width=\textwidth]{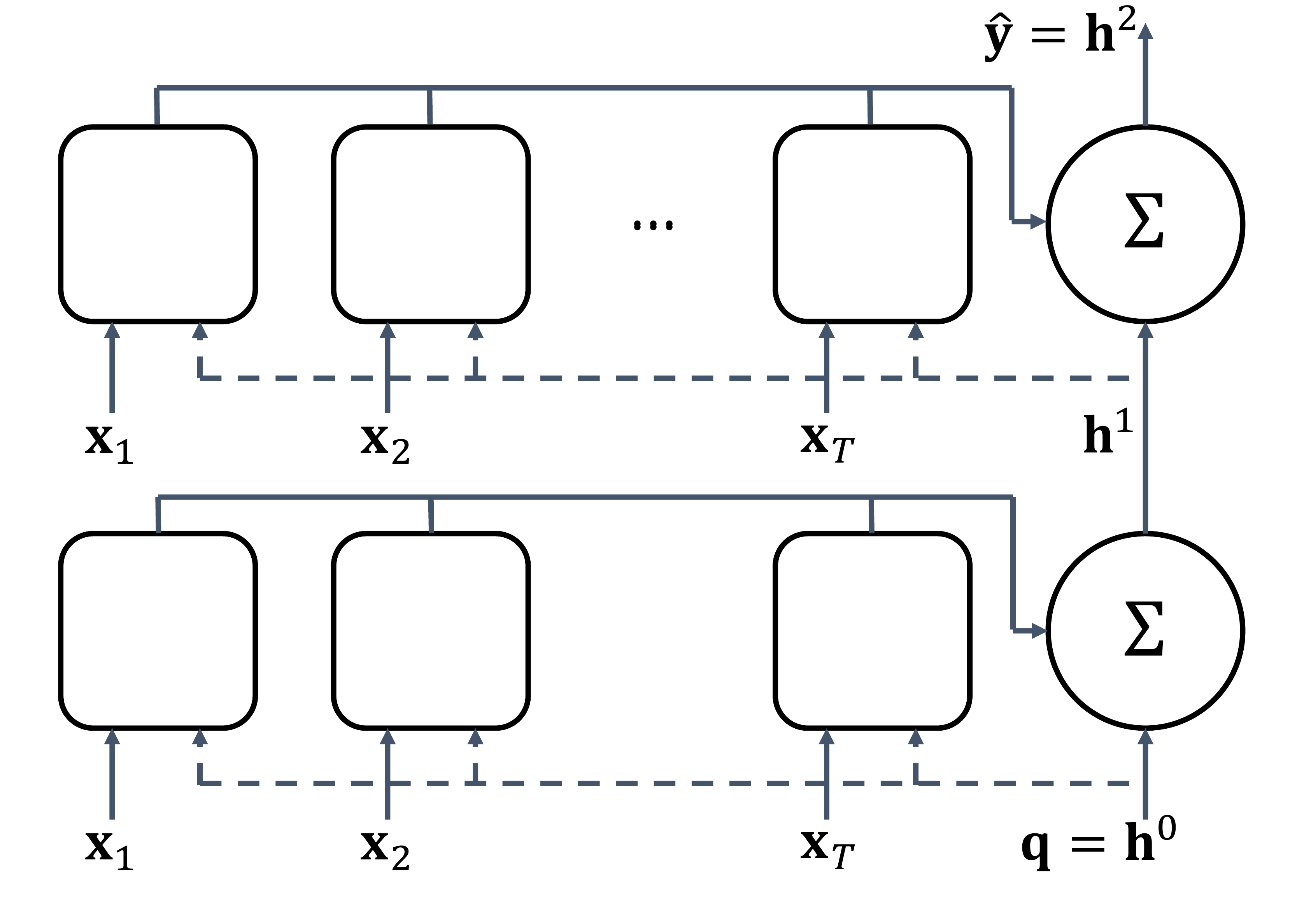}
\caption{ N2N~\citep{memN2N} }
\label{fig:memn2n}
\end{subfigure}
\begin{subfigure}[htbp]{0.32\textwidth}
\centering
\includegraphics[width=\textwidth]{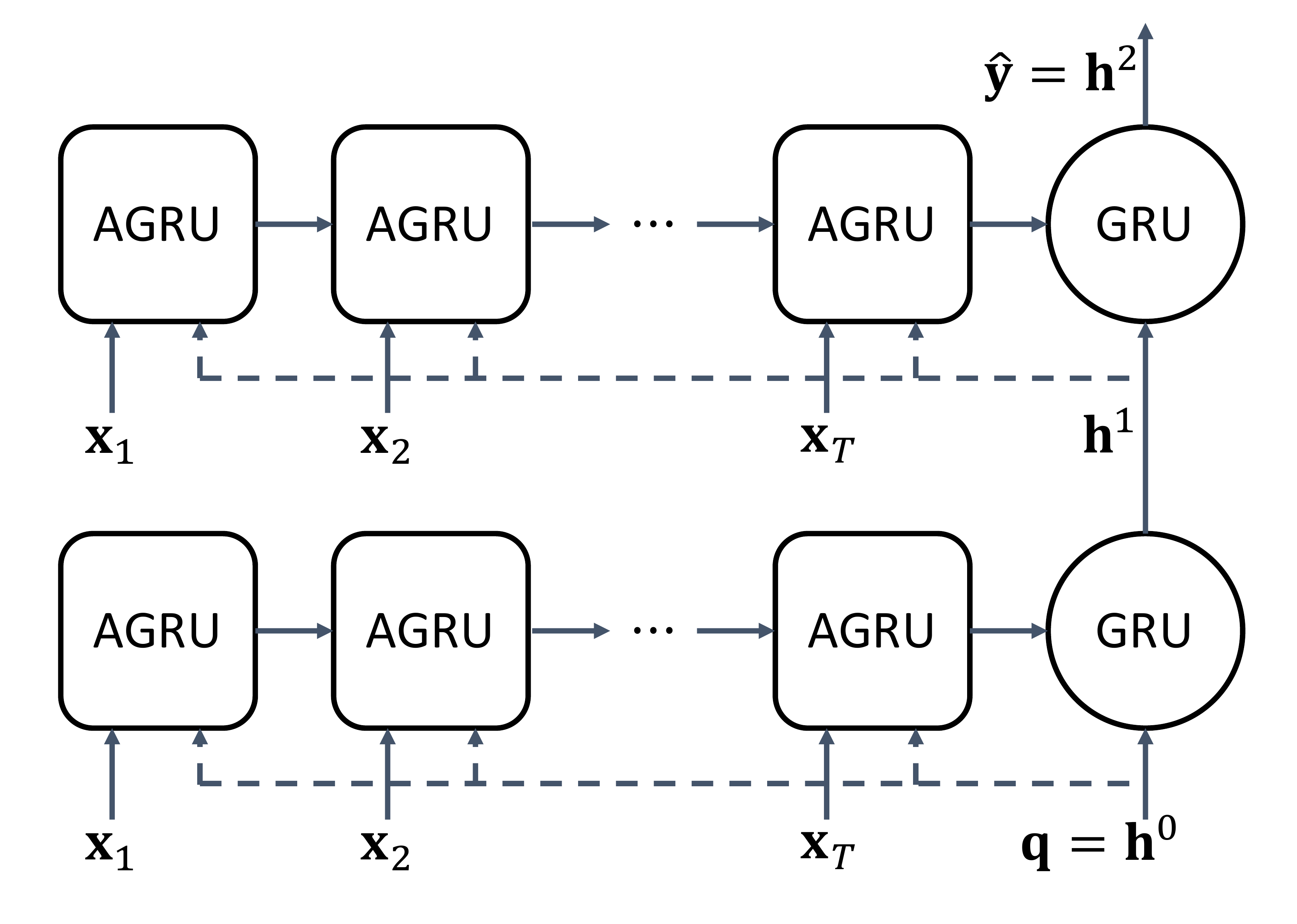}
\caption{ DMN+~\citep{DMN+} }
\label{fig:dmn}
\end{subfigure}
\caption{\small The schematics of QRN and the two state-of-the-art models, End-to-End Memory Networks (N2N) and Improved Dynamic Memory Networks (DMN+), simplified to emphasize the differences among the models. 
AGRU is a variant of GRU where the update gate is replaced with soft attention, proposed by \citet{DMN}.
For QRN and DMN+, only forward direction arrows are shown.}
\label{fig:models}
\end{figure}

QRN is inspired by RNN-based models with gating mechanism, such as LSTM~\citep{lstm} and GRU~\citep{GRU}.
While GRU and LSTM use the previous hidden state and the current input to obtain the candidate hidden state, QRN only uses the current two inputs to obtain the candidate reduced query (equivalent to candidate hidden state).
We conjecture that this not only gives computational advantage via parallelization, but also makes training easier, i.e., avoiding vanishing gradient (which is critical for long-term dependency), overfitting (by simplifying the model), and converging to local minima.

The idea of structurally simplifying (constraining) RNNs for learning longer-term patterns has been explored in recent previous work, such as Structurally Constrained Recurrent Network~\citep{scrn} and Strongly-Typed Recurrent Neural Network (STRNN)~\citep{strnn}. 
QRN is similar to STRNN in that both architectures use gating mechanism, and the gates and the candidate hidden states do not depend on the previous hidden states, which simplifies the recurrent relation. 
However, QRN can be distinguished from STRNN in three ways.
First, QRN’s update gate simulates attention mechanism, measuring the relevance between the input sentence and query. On the other hand, the gates in STRNN can be considered as the simplification of LSTM/GRU by removing their dependency on previous hidden state.
Second, QRN is an RNN that is natively compatible with context-based QA tasks, where the QRN unit accepts two inputs, i.e. each context sentence and query. 
This is distinct from STRNN which has only one input. 
Third, we show that QRN is timewise-parallelizable on GPUs. Our parallelization algorithm is also applicable to STRNN.

End-to-end Memory Network (N2N)~\citep{memN2N} uses external memory with multi-layer attention mechanism to focus on sentences that are relevant to the question.
There are two key differences between N2N and our  QRN. 
First, N2N summarizes the entire memory in each layer to control the attention in the next layer (circle nodes in Figure~\ref{fig:memn2n}).
Instead, QRN does not have any controller node (Figure~\ref{fig:qrn}) and is able to focus on relevant sentences through the update gate that is internally embodied within its unit.
Second, N2N adds time-dependent trainable weights to the sentence representations to model the time dependency of the sentences (as discussed in Section~\ref{sec:intro}).
QRN does not need such additional weights as its inherent RNN architecture allows QRN to effectively model the time dependency.
Neural Reasoner~\citep{NR} and Gated End-to-end Memory Network~\citep{perez2016gated}) are variants of MemN2N that share its fundamental characteristics.

Improved Dynamic Memory Network (DMN+)~\citep{DMN+} uses the hybrid of the attention mechanism and the RNN architecture to model the sequence of sentences.
It consists of two distinct GRUs, one for the time axis (rectangle nodes in Figure~\ref{fig:dmn}) and one for the layer axis (circle nodes in Figure~\ref{fig:dmn}).
Note that the update gate of the GRU for the time axis is replaced with external softmax attention weights.
DMN+ uses the time-axis GRU to summarizes the entire memory in each layer, and then the layer-axis GRU controls the attention weights in each layer.
In contrast, QRN is simply a single recurrent unit without any controller node.

\section{Experiments}\label{sec:qa}

\subsection{Data}
\paragraph{bAbI story-based QA dataset} bAbI story-based QA dataset~\citep{babi} is composed of 20 different tasks (Appendix~\ref{sec:app-a}), each of which has 1,000 (1k) synthetically-generated story-question pair. 
A story can be as short as two sentences and as long as 200+ sentences.
A system is evaluated on the accuracy of  getting the correct answers to the questions. 
The answers are single words or lists (e.g. ``football, apple'').
Answering questions in each task requires selecting a set of relevant sentences and applying different kinds of logical reasoning over them. 
The dataset also includes 10k training data (for each task), which allows training more complex models. 
Note that DMN+~\citep{DMN+} only reports on the 10k dataset. 
\paragraph{bAbI dialog dataset}
bAbI dialog dataset~\citep{bordes2016learning} consists of 5 different tasks (Table~\ref{tab:dialog-all}), each of which has 1k synthetically-generated goal-oriented dialogs between a user and the system in the domain of restaurant reservation. 
Each dialog is as long as 96 utterances and comes with external knowledge base (KB) providing information of each restaurant.
The authors also provide Out-Of-Vocabulary (OOV) version of the dataset, where many of the words and KB keywords in test data are not seen during training.
A system is evaluated on the accuracy of its response to each utterance of the user, choosing from up to 2500 possible candidate responses.
A system is required not only to understand the user's request but also refer to previous conversations in order to obtain the context information of the current conversation.


\paragraph{DSTC2 (Task 6) dialog dataset} 
~\cite{bordes2016learning} transformed the Second Dialog State Tracking Challenge (DSTC2) dataset~\citep{henderson2014second} into the same format as the bAbI dialog dataset, for the measurement of performance on a real dataset.
Each dialog can be as long as 800+ utterances, and a system needs to choose from 2407 possible candidate responses for each utterance of the user.
Note that the evaluation metric of the original DSTC2 is different from that of the transformed DSTC2, so previous work on the original DSTC2 should not be directly compared to our work.
We will refer to this transformed DSTC2 dataset by ``Task 6'' of dialog dataset.


\subsection{Model Details}

\paragraph{Input Module.} 
In the input module, we are given sentences (previous conversations in dialog) ${\bm x}_t$ and a question (most recent user utterance) ${\bm q}$, and we want to obtain their vector representations, ${\bf x}_t, {\bf q} \in \mathbb{R}^d$.
We use a trainable embedding matrix ${\bf A} \in \mathbb{R}^{d \times V}$ to encode the one-hot vector of each word ${\bm x}_{tj}$ in each sentence ${\bm x}_t$ into a $d$-dimensional vector ${\bf x}_{tj} \in \mathbb{R}^d$.
Then the sentence representation ${\bf x}_t$ is obtained by Position Encoder~\citep{memNet}. 
The same encoder with the same embedding matrix is also used to obtain the question vector ${\bf q}$ from ${\bm q}$.

\paragraph{Output Module for story-based QA.} 
In the output module, we are given the vector representation of the predicted answer $\hat{\bf y}$ and we want to obtain the natural language form of the answer, $\hat{\bm y}$.
We use a $V$-way single-layer softmax classifier to map $\hat{\bf y}$ to a $V$-dimensional sparse vector, $\hat{\bf v} = \mathrm{softmax}\left( {\bf W}^{(y)} \hat{\bf y} \right) \in \mathbb{R}^V$, where ${\bf W}^{(y)} \in \mathbb{R}^{V \times d}$ is a weight matrix.
Then the final answer $\hat{\bm y}$ is simply the argmax word in $\hat{\bf v}$. 
To handle questions with multiple-word answers, we consider each of them as a single word that contains punctuations such as space and comma, and put it in the vocabulary.

\paragraph{Output Module for dialog.}
We use a fixed number single-layer softmax classifiers, each of which is similar to that of the sotry-based QA model, to sequentially output each word of the system's response.
While it is similar in spirit to the RNN decoder~\citep{GRU}, our output module does not have a recurrent hidden state or gating mechanism.
Instead, it solely uses the final ouptut of the QRN, $\hat{\bf y}$, and the current word output to influence the prediction of the next word among possible candidates.


\paragraph{Training.} 
We withhold $10\%$ of the training  for development.
We use the hidden state size of 50 by deafult.
Batch sizes of $32$ for bAbI story-based QA 1k, bAbI dialog and DSTC2 dialog, and $128$ for bAbI QA 10k are used.
The weights in the input and output modules are initialized with zero mean and the standard deviation  of $1/\sqrt{d}$.
Weights in the QRN unit are initialized using techniques by \citet{glorot2010understanding}, and are tied across the layers.
Forget bias of $2.5$ is used for update gates (no bias for reset gates).
L2 weight decay of $0.001$ ($0.0005$ for QA 10k) is used for all weights.
The loss function is the cross entropy between $\hat{\bf v}$ and the one-hot vector of the true answer.
The loss is minimized by stochastic gradient descent for maximally $500$ epochs, but training is early stopped if the loss on the development data does not decrease for $50$ epochs.
The learning rate is controlled by AdaGrad~\citep{duchi2011adaptive}
with the initial learning rate of $0.5$ ($0.1$ for QA 10k).
Since the model is sensitive to the weight initialization,
we repeat each training procedure 10 times (50 times for 10k) with the new random initialization of the weights and report the result on the test data with the lowest loss on the development data.

\subsection{Results.}\label{subsec:results}
We compare our model with baselines and previous state-of-the-art models on story-based and dialog tasks (Table~\ref{tab:summary}). 
These include LSTM~\citep{lstm}, End-to-end Memory Networks (N2N)~\citep{memN2N}, Dynamic Memory Networks (DMN+)~\citep{DMN+}, Gated End-to-end Memory Networks (GMemN2N)~\citep{perez2016gated}, and Differentiable Neural Computer (DNC)~\citep{graves2016hybrid}. 

\paragraph{Story-based QA.} Table~\ref{tab:summary}(top) reports the summary of results of our model (QRN) and previous work on bAbI QA
(task-wise results are shown in Table~\ref{tab:qa-all} in Appendix).
In 1k data, QRN's `2r' (2 layers + reset gate + $d=50$) outperforms all other models by a large margin (2.8+\%).
In 10k dataset, the average accuracy of QRN's `6r200' (6 layers + reset gate + $d=200$) model outperforms all previous models by a large margin (2.5+\%), achieving a nearly perfect score of 99.7\%.

\paragraph{Dialog.}
 Table~\ref{tab:summary}(bottom) reports the summary of the results of our model (QRN) and previous work on bAbI dialog and Task 6 dialog (task-wise results are shown in Table~\ref{tab:dialog-all} in Appendix).
As done in previous work~\citep{bordes2016learning, perez2016gated}, we also report results when we use `Match' for dialogs. `Match' is the extension to the model which additionally takes as input whether each answer candidate matches with context (more details on Appendix).  QRN outperforms  previous work by a large margin (2.0+\%) in every comparison.


\begin{table}
\centering
\resizebox{\columnwidth}{!}{
\begin{tabular}{l||c|c|c|c|c|c||c|c|c|c|c} 
 \hline
\multirow{3}{*}{Task} & \multicolumn{6}{|c||}{1k} & \multicolumn{5}{|c}{10k} \\
\cline{2-12}
 & \multicolumn{4}{|c|}{Previous works} & \multicolumn{2}{|c||}{QRN} & \multicolumn{4}{|c|}{Previous works} & \multicolumn{1}{|c}{QRN}\\
 \cline{2-12}
                        & LSTM  & N2N   &DMN+$^\dagger$   & GMemN2N &2r    &3r   & N2N   & DMN+      & GMemN2N & DNC &6r200 \\ 
 \hline
  \# Failed             & 20    & 10    & 16    & 10    & 7     &{\bf 5} & 3     & 1   & 3       & 2   &{\bf 0}   \\ 
 Average error rates    & 51.3  & 15.2  & 33.2  & 12.7  &{\bf 9.9}& 11.3 & 4.2   & 2.8 & 3.7     & 3.8 &{\bf 0.3} \\
 \hline \hline 
\end{tabular}
} 
\\
\resizebox{0.88\columnwidth}{!}{
\begin{tabular}{l||c|c|c|c||c|c|c} 
 \hline
\multirow{3}{*}{Task} & \multicolumn{4}{|c||}{Plain} & \multicolumn{3}{|c}{With Match} \\
\cline{2-8} & \multicolumn{2}{|c|}{Previous works} & \multicolumn{2}{|c||}{QRN} & \multicolumn{2}{|c|}{Previous works} & \multicolumn{1}{|c}{QRN} \\
\cline{2-8}
                                      & N2N & GMemN2N &2r   & 2r100    & N2N+ & GMemN2N+ & 2r+      \\ 
 \hline

bAbI dialog Average error rates       & 13.9  & 14.3  &{\bf 5.5}  & {\bf 5.5} & 6.7   & 5.4   &{\bf 1.5}  \\
bAbI dialog (OOV) Average error rates & 30.3  & 27.9  &{\bf 11.1} &{\bf 11.1} & 11.2  & 10.3  &{\bf 2.3} \\
 \hline
DSTC2 dialog Average error rates      & 58.9  & 52.6  & 49.5      & {\bf 48.9}   & 59.0  & 51.3  &{\bf 49.3}  \\ 
 
 \hline 
\end{tabular}
}

\caption{\small (top) bAbI QA dataset~\citep{babi}: number of failed tasks and average error rates (\%).
$^\dagger$ is obtained from \url{github.com/therne/dmn-tensorflow}.
(bottom) bAbI dialog and DSTC2 dialog dataset~\citep{bordes2016learning} average error rates (\%) of QRN and previous work (LSTM, N2N, DMN+, GMemN2N,  and DNC).  
For QRN, the first number (1, 2, 3) indicates the number of layers, `r' means the reset gate is used, and the last number (100, 200), if exists, indicates the dimension of the hidden state, where the default value is 50. `+' indicates that `match' (See Appendix for details) is used. The task-wise results are shown in Appendices: Table~\ref{tab:qa-all} (bAbI QA) and Table~\ref{tab:dialog-all} (dialog datasets). See Section~\ref{subsec:results} for details.}
\label{tab:summary}
\end{table}


\paragraph{Ablations.}
We test four types of ablations (also discussed in Section~\ref{sec:var}): number of layers (1, 2, 3, or 6), reset gate (r), and gate vectorization (v) and the dimension of the hidden vector (50, 100). 
We show a subset of combinations of the ablations for bAbI QA in Table~\ref{tab:summary} and Table~\ref{tab:qa-all}; other combinations performed poorly and/or did not give interesting observations.
According to the ablation results, we  infer that:
{\bf (a)} When the number of layers is only one, the model lacks reasoning capability. In the case of 1k dataset, when there are too many layers (6), it seems correctly training the model becomes increasingly difficult. In the case of 10k dataset, many layers (6) and hidden dimensions (200) helps reasoning, most notably in difficult task such as task 16.
{\bf (b)} Adding the reset gate helps.
{\bf (c)} Including vector gates hurts in 1k datasets, as the model either overfits to the training data or converges to local minima. On the other hand, vector gates in bAbI story-based QA 10k dataset sometimes help.
{\bf (d)} Increasing the dimension of the hidden state to 100 in the dialog's Task 6 (DSTC2) helps, while there is not much improvement in the dialog's Task 1-5. 
It can be hypothesized that a larger hidden state is required for real data.



\begin{figure}[t]
    \centering
    \resizebox{\columnwidth}{!}{
    \begin{tabular}{|l|ccc|c|}
    \hline
    & \multicolumn{3}{|c|}{Layer 1} & Layer 2\\
    \hline
    Task 2: Two Supporting Facts & $z^1$ & $\overrightarrow{r}^1$ & $\overleftarrow{r}^1$ & $z^2$\\
    \hline
    Sandra picked up the apple there. & \cellcolor{red!95} 0.95 & \cellcolor{red!89}0.89 & \cellcolor{red!98}0.98 & 0.00\\
    Sandra dropped the apple. & \cellcolor{red!83}0.83 & \cellcolor{red!5} 0.05 & \cellcolor{red!92}0.92 & \cellcolor{red!1} 0.01\\
    Daniel grabbed the apple there. & \cellcolor{red!88} 0.88 & \cellcolor{red!93} 0.93 & \cellcolor{red!98} 0.98 & 0.00\\
    Sandra travelled to the bathroom. & \cellcolor{red!1} \cellcolor{red!1}0.01 & \cellcolor{red!18}0.18 & \cellcolor{red!63}0.63 & \cellcolor{red!2}0.02\\
    Daniel went to the hallway. & \cellcolor{red!1}0.01 & \cellcolor{red!24}0.24 & \cellcolor{red!62}0.62 & \cellcolor{red!83}0.83\\
    \hline
    \multicolumn{5}{|l|}{Where is the apple? \hfill hallway}\\
    \hline
    \end{tabular}
    \begin{tabular}{|l|ccc|c|}
    \hline
    & \multicolumn{3}{|c|}{Layer 1} & Layer 2\\
    \hline
    Task 15: Deduction & $z^1$ & $\overrightarrow{r}^1$ & $\overleftarrow{r}^1$ & $z^2$\\
    \hline
    Mice are afraid of wolves. & \cellcolor{red!11} 0.11 & \cellcolor{red!99}0.99 & \cellcolor{red!13}0.13 & \cellcolor{red!78}0.78\\
    Gertrude is a mouse. & \cellcolor{red!77}0.77 & \cellcolor{red!99}0.99 & \cellcolor{red!96}0.96 & \cellcolor{red!00}0.00\\
    Cats are afraid of sheep. & \cellcolor{red!1}0.01 & \cellcolor{red!99}0.99 & \cellcolor{red!7}0.07 & \cellcolor{red!03}0.03\\
    Winona is a mouse. & \cellcolor{red!14}0.14 & \cellcolor{red!85}0.85 & \cellcolor{red!77}0.77 & \cellcolor{red!05}0.05\\
    Sheep are afraid of wolves. & \cellcolor{red!02}0.02 & \cellcolor{red!98}0.98 & \cellcolor{red!27}0.27 & \cellcolor{red!05}0.05\\
    \hline
    \multicolumn{5}{|l|}{What is Gertrude afraid of? \hfill wolf}\\
    \hline
    \end{tabular}
    }
    
   \resizebox{\columnwidth}{!}{
    \begin{tabular}{|p{4.5cm}|ccc|c|}
    \hline
    & \multicolumn{3}{|c|}{Layer 1} & Layer 2\\
    \hline
    Task 3: Displaying options & $z^1$ & $\overrightarrow{r}^1$ & $\overleftarrow{r}^1$ & $z^2$\\
    \hline
    
    resto-paris-expen-frech-8stars? & \cellcolor{red!0} 0.00 & \cellcolor{red!100}1.00 & \cellcolor{red!96}0.96 & \cellcolor{red!91}0.91\\
    Do you have something else? & \cellcolor{red!41}0.41 & \cellcolor{red!99}0.99 & \cellcolor{red!00}0.00 & \cellcolor{red!00}0.00\\
   Sure let me find another option. & \cellcolor{red!100}1.00 & \cellcolor{red!0}0.00 & \cellcolor{red!0}0.00 & \cellcolor{red!12}0.12\\
     resto-paris-expen-frech-5stars? & \cellcolor{red!0}0.00 & \cellcolor{red!100}1.00 & \cellcolor{red!96}0.96 & 
    \cellcolor{red!91}0.91\\
    No this does not work for me. & \cellcolor{red!00}0.00 & \cellcolor{red!00}0.00 & \cellcolor{red!14}0.14 & \cellcolor{red!00}0.00\\
    Sure let me find an other option.  & \cellcolor{red!100}1.00 & \cellcolor{red!0}0.00 & \cellcolor{red!0}0.00 & \cellcolor{red!12}0.12\\
    
    \hline

    \multicolumn{5}{|l|}{
   \hfill What do you think of this? resto-paris-expen-french-4stars} \\
    \hline
    \end{tabular}
    
    \begin{tabular}{|p{6.5cm}|ccc|c|}
    \hline
    & \multicolumn{3}{|c|}{Layer 1} & Layer 2\\
    \hline
    Task 6: DSTC2 dialog & $z^1$ & $\overrightarrow{r}^1$ & $\overleftarrow{r}^1$ & $z^2$\\
    \hline
    
    Spanish food. & \cellcolor{red!84}0.84 & \cellcolor{red!7}0.07 & \cellcolor{red!0}0.00 & \cellcolor{red!82}0.82\\
    You are lookng for a spanish restaurant right? & \cellcolor{red!98}0.98 & \cellcolor{red!2}0.02 & \cellcolor{red!49}0.49 & \cellcolor{red!75}0.75\\
    Yes. & \cellcolor{red!1}0.01 & \cellcolor{red!100}1.00 & \cellcolor{red!33}0.33 & \cellcolor{red!13}0.13\\
    What part of town do you have in mind? & \cellcolor{red!20}0.20 & \cellcolor{red!73}0.73 & \cellcolor{red!41}0.41 & 
    \cellcolor{red!11}0.11\\
    I don't care. & \cellcolor{red!00}0.00 & \cellcolor{red!100}1.00 & \cellcolor{red!2}0.02 & \cellcolor{red!00}0.00\\
    What price range would you like?  & \cellcolor{red!72}0.72 & \cellcolor{red!46}0.46 & \cellcolor{red!52}0.52 & \cellcolor{red!72}0.72\\
    
    \hline
    
    \multicolumn{5}{|l|}{I don't care. \hfill API CALL spanish R-location R-price}\\
    
    \hline
    \end{tabular}
    }
    
    \caption{\small (top) bAbI QA dataset~\citep{babi} visualization of update and reset gates in QRN `2r' model (bottom two) bAbI dialog and DSTC2 dialog dataset~\citep{bordes2016learning} visualization of update and reset gates in QRN `2r' model.
    Note that the stories can have as many as 800+ sentences; we only show part of them here. More visualizations are shown in Figure~\ref{fig:qa-att-all}~(bAbI QA) and Figure~\ref{fig:dialog-att-all}~(dialog datasets).
    }
    \label{fig:att-summary}
\end{figure}


\paragraph{Parallelization.}
We implement QRN with and without parallelization in TensorFlow~\citep{tensorflow} on a single Titan X GPU to qunaitify the computational gain of the parallelization.
For QRN without parallelization, we use the RNN library provided by TensorFlow. 
QRN with parallelization gives 6.2 times faster training and inference than QRN without parallelization on average.
We expect that the speedup can be even higher for datasets with larger context.

\paragraph{Interpretations.}
An advantage of QRN is that the intermediate query updates are interpretable.
Figure~\ref{fig:model} shows intermediate local queries (${\bf q}^k_t$) interpreted in natural language, such as ``Where is Sandra?''.
In order to obtain these, we place a decoder on the input question embedding ${\bf q}$ and add its loss for recovering the question to the classification loss (similarly to \citet{NR}).
We then use the same decoder to decode the intermediate queries.
This helps us understand the flow of information in the networks. 
In Figure~\ref{fig:model}, the question \texttt{Where is apple?} is transformed into \texttt{Where is Sandra?} at $t=1$.
At $t=2$, as \texttt{Sandra dropped the apple}, the apple is no more relevant to Sandra.
We obtain \texttt{Where is Daniel?} at time $t = 3$, and it is propagated until $t=5$, where we observe a sentence (fact) that can be used to answer the query.

\paragraph{Visualization.}
Figure~\ref{fig:att-summary} shows vizualization of the (scalar) magnitudes of update and reset gates on story sentences and dialog utterances.
More visualizations are shown in Appendices: Figure~\ref{fig:qa-att-all} and Figure~\ref{fig:dialog-att-all}.
In Figure~\ref{fig:att-summary}, we observe high values on facts that provide information to answer question (the system's next utterance for dialog).
In QA Task 2 example (top left), we observe high update gate values  in the first layer on facts that state who has the \texttt{apple}, and in the second layer, the high update gate values are on those that inform where that person went to. We also observe that the forward reset gate at $t=2$ in the first layer ($\overrightarrow{r}^1_2$) is low, which is signifying that \texttt{apple} no more belongs to \texttt{Sandra}. 
In dialog Task 3~(bottom left), the model is able to infer that three restaurants are already recommended so that it can recommend another one. In dialog Task 6~(bottom), the model focuses on the sentences containing \texttt{Spanish}, and does not concentrate much on other facts such as \texttt{I don't care}.


\section{Conclusion} 
\vspace{-.3cm}
In this paper, we introduce Query-Reduction Network (QRN)  to answer context-based questions and carry out conversations with users that require multi-hop reasoning. 
We show the state-of-the-art results in the three datasets of story-based QA and dialog. 
We model a story or a dialog as a sequence of state-changing triggers and compute the final answer to the question or the system's next utterance by recurrently updating (or \emph{reducing}) the query.
QRN is situated between the attention mechanism and RNN, effectively handling time dependency and long-term dependency problems of each technique, respectively.  
It addresses the long-term dependency problem of most RNNs by simplifying the recurrent update, in which the candidate hidden state (reduced query) does not depend on the previous state. Moreover, QRN can be parallelized and can address the well-known problem of RNN's vanishing gradients.
\subsubsection*{Acknowledgments}
This research was supported by the NSF (IIS 1616112), Allen Institute for AI (66-9175), Allen Distinguished Investigator Award, Google Research Faculty Award, and Samsung GRO Award. We thank the anonymous reviewers for their helpful comments.

\bibliographystyle{plainnat}
\bibliography{00-main}

\begin{thebibliography}{22}
\providecommand{\natexlab}[1]{#1}
\providecommand{\url}[1]{\texttt{#1}}
\expandafter\ifx\csname urlstyle\endcsname\relax
  \providecommand{\doi}[1]{doi: #1}\else
  \providecommand{\doi}{doi: \begingroup \urlstyle{rm}\Url}\fi

\bibitem[Abadi et~al.(2016)Abadi, Agarwal, Barham, Brevdo, Chen, Citro,
  Corrado, Davis, Dean, Devin, et~al.]{tensorflow}
Mart{\i}n Abadi, Ashish Agarwal, Paul Barham, Eugene Brevdo, Zhifeng Chen,
  Craig Citro, Greg~S Corrado, Andy Davis, Jeffrey Dean, Matthieu Devin, et~al.
\newblock Tensorflow: Large-scale machine learning on heterogeneous distributed
  systems.
\newblock \emph{arXiv preprint arXiv:1603.04467}, 2016.

\bibitem[Balduzzi and Ghifary(2016)]{strnn}
David Balduzzi and Muhammad Ghifary.
\newblock Strongly-typed recurrent neural networks.
\newblock In \emph{ICML}, 2016.

\bibitem[Bordes and Weston(2016)]{bordes2016learning}
Antoine Bordes and Jason Weston.
\newblock Learning end-to-end goal-oriented dialog.
\newblock \emph{arXiv preprint arXiv:1605.07683}, 2016.

\bibitem[Cho et~al.(2014)Cho, Van~Merri{\"e}nboer, Gulcehre, Bahdanau,
  Bougares, Schwenk, and Bengio]{GRU}
Kyunghyun Cho, Bart Van~Merri{\"e}nboer, Caglar Gulcehre, Dzmitry Bahdanau,
  Fethi Bougares, Holger Schwenk, and Yoshua Bengio.
\newblock Learning phrase representations using rnn encoder-decoder for
  statistical machine translation.
\newblock In \emph{EMNLP}, 2014.

\bibitem[Duchi et~al.(2011)Duchi, Hazan, and Singer]{duchi2011adaptive}
John Duchi, Elad Hazan, and Yoram Singer.
\newblock Adaptive subgradient methods for online learning and stochastic
  optimization.
\newblock \emph{JMLR}, 12, 2011.

\bibitem[Glorot and Bengio(2010)]{glorot2010understanding}
Xavier Glorot and Yoshua Bengio.
\newblock Understanding the difficulty of training deep feedforward neural
  networks.
\newblock In \emph{JMLR}, 2010.

\bibitem[Graves et~al.(2016)Graves, Wayne, Reynolds, Harley, Danihelka,
  Grabska-Barwi{\'n}ska, Colmenarejo, Grefenstette, Ramalho, Agapiou,
  et~al.]{graves2016hybrid}
Alex Graves, Greg Wayne, Malcolm Reynolds, Tim Harley, Ivo Danihelka, Agnieszka
  Grabska-Barwi{\'n}ska, Sergio~G{\'o}mez Colmenarejo, Edward Grefenstette,
  Tiago Ramalho, John Agapiou, et~al.
\newblock Hybrid computing using a neural network with dynamic external memory.
\newblock \emph{Nature}, 2016.

\bibitem[Henderson et~al.(2014)Henderson, Thomson, and
  Williams]{henderson2014second}
Matthew Henderson, Blaise Thomson, and Jason Williams.
\newblock The second dialog state tracking challenge.
\newblock In \emph{SIGdial}, 2014.

\bibitem[Hermann et~al.(2015)Hermann, Kocisky, Grefenstette, Espeholt, Kay,
  Suleyman, and Blunsom]{hermann2015teaching}
Karl~Moritz Hermann, Tomas Kocisky, Edward Grefenstette, Lasse Espeholt, Will
  Kay, Mustafa Suleyman, and Phil Blunsom.
\newblock Teaching machines to read and comprehend.
\newblock In \emph{NIPS}, 2015.

\bibitem[Hill et~al.(2016)Hill, Bordes, Chopra, and Weston]{hill2015goldilocks}
Felix Hill, Antoine Bordes, Sumit Chopra, and Jason Weston.
\newblock The goldilocks principle: Reading children's books with explicit
  memory representations.
\newblock In \emph{ICLR}, 2016.

\bibitem[Hochreiter and Schmidhuber(1997)]{lstm}
Sepp Hochreiter and J{\"u}rgen Schmidhuber.
\newblock Long short-term memory.
\newblock \emph{Neural computation}, 9\penalty0 (8):\penalty0 1735--1780, 1997.

\bibitem[Kumar et~al.(2016)Kumar, Irsoy, Su, Bradbury, English, Pierce,
  Ondruska, Gulrajani, and Socher]{DMN}
Ankit Kumar, Ozan Irsoy, Jonathan Su, James Bradbury, Robert English, Brian
  Pierce, Peter Ondruska, Ishaan Gulrajani, and Richard Socher.
\newblock Ask me anything: Dynamic memory networks for natural language
  processing.
\newblock In \emph{ICML}, 2016.

\bibitem[Mikolov et~al.(2015)Mikolov, Joulin, Chopra, Mathieu, and
  Ranzato]{scrn}
Tomas Mikolov, Armand Joulin, Sumit Chopra, Michael Mathieu, and Marc'Aurelio
  Ranzato.
\newblock Learning longer memory in recurrent neural networks.
\newblock In \emph{ICLR 2015 Workshop}, 2015.

\bibitem[Peng et~al.(2015)Peng, Lu, Li, and Wong]{NR}
Baolin Peng, Zhengdong Lu, Hang Li, and Kam-Fai Wong.
\newblock Towards neural network-based reasoning.
\newblock \emph{arXiv preprint arXiv:1508.05508}, 2015.

\bibitem[Perez and Liu(2016)]{perez2016gated}
Julien Perez and Fei Liu.
\newblock Gated end-to-end memory networks.
\newblock \emph{arXiv preprint arXiv:1610.04211}, 2016.

\bibitem[Rajpurkar et~al.(2016)Rajpurkar, Zhang, Lopyrev, and Liang]{squad}
Pranav Rajpurkar, Jian Zhang, Konstantin Lopyrev, and Percy Liang.
\newblock Squad: 100,000+ questions for machine comprehension of text.
\newblock In \emph{EMNLP}, 2016.

\bibitem[Reiter(2001)]{reiter}
Raymond Reiter.
\newblock \emph{Knowledge in Action}.
\newblock MIT Press, 1st edition, 2001.

\bibitem[Richardson et~al.(2013)Richardson, Burges, and Renshaw]{MCTest}
Matthew Richardson, Christopher~JC Burges, and Erin Renshaw.
\newblock Mctest: A challenge dataset for the open-domain machine comprehension
  of text.
\newblock In \emph{EMNLP}, 2013.

\bibitem[Sukhbaatar et~al.(2015)Sukhbaatar, Szlam, Weston, and Fergus]{memN2N}
Sainbayar Sukhbaatar, Arthur Szlam, Jason Weston, and Rob Fergus.
\newblock End-to-end memory networks.
\newblock In \emph{NIPS}, 2015.

\bibitem[Weston et~al.(2015)Weston, Chopra, and Bordes]{memNet}
Jason Weston, Sumit Chopra, and Antoine Bordes.
\newblock Memory networks.
\newblock In \emph{ICLR}, 2015.

\bibitem[Weston et~al.(2016)Weston, Bordes, Chopra, and Mikolov]{babi}
Jason Weston, Antoine Bordes, Sumit Chopra, and Tomas Mikolov.
\newblock Towards ai-complete question answering: A set of prerequisite toy
  tasks.
\newblock In \emph{ICLR}, 2016.

\bibitem[Xiong et~al.(2016)Xiong, Merity, and Socher]{DMN+}
Caiming Xiong, Stephen Merity, and Richard Socher.
\newblock Dynamic memory networks for visual and textual question answering.
\newblock In \emph{ICML}, 2016.

\end{thebibliography}

\newpage
\appendix
\section{Task-wise Results}\label{sec:app-a}
Here we provide detailed per-task breakdown of our results in QA(Table~\ref{tab:qa-all}) and dialog datasets (Table~\ref{tab:dialog-all}).  
\\

\begin{table}[ht]
\begin{center}
\resizebox{\columnwidth}{!}{
\begin{tabular}{|l||c|c|c|c|c|c|c|c|c|c||c|c|c|c|c|c|c|} 
 \hline
\multirow{3}{*}{Task} & \multicolumn{10}{|c||}{1k} & \multicolumn{7}{|c|}{10k} \\
\cline{2-18}
 & \multicolumn{4}{|c|}{Previous works} & \multicolumn{6}{|c||}{QRN} & \multicolumn{3}{|c|}{Previous works} & \multicolumn{4}{|c|}{QRN}\\
 \cline{2-18}
                                & LSTM  & N2N   &DMN+   &GMemN2N    & 1r   & 2    & 2r   & 3r   & 6r   & 6r200* & N2N   & DMN+  &GMemN2N & 2r   & 2rv  & 3r & 6r200\\ 
 \hline\textbf{}
 1: Single supporting fact      & 50.0  & 0.1   & 1.3   & 0.0   & 0.0   & 0.0   & 0.0   & 0.0   & 0.0   &13.1   & 0.0   & 0.0   & 0.0   & 0.0   & 0.0   & 0.0    &0.0\\ 
 2: Two supporting facts        & 80.0  & 18.8  & 72.3  & 8.1   & 65.7  & 1.2   & 0.7   & 0.5   & 1.5   &15.3   & 0.3   & 0.3   & 0.0   & 0.4   & 0.8   & 0.4    &0.0\\
 3: Three supporting facts      & 80.0  & 31.7  & 73.3  & 38.7  & 68.2  & 17.5  & 5.7   & 1.2   & 15.3  &13.8   & 2.1   & 1.1   & 4.5   & 0.4   & 1.4   & 0.0    &0.0\\
 4: Two arg relations           & 39.0  & 17.5  & 26.9  & 0.4   & 0.0   & 0.0   & 0.0   & 0.7   & 9.0   &13.6   & 0.0   & 0.0   & 0.0   & 0.0   & 0.0   & 0.0    &0.0\\
 5: Three arg relations         & 30.0  & 12.9  & 25.6  & 1.0   & 1.0   & 1.1   & 1.1   & 1.2   & 1.3   &12.5   & 0.8   & 0.5   & 0.2   & 0.5   & 0.2   & 0.3    &0.0\\ 
 6: Yes/no questions            & 52.0  & 2.0   & 28.5  & 8.4   & 0.1   & 0.0   & 0.9   & 1.2   & 50.6  &15.5   & 0.1   & 0.0   & 0.0   & 0.0   & 0.0   & 0.0    &0.0\\
 7: Counting                    & 51.0  & 10.1  & 21.9  & 17.8  & 10.9  & 11.1  & 9.6   & 9.4   & 13.1  &15.3   & 2.0   & 2.4   & 1.8   & 1.0   & 0.7   & 0.7    &0.0\\
 8: Lists/sets                  & 55.0  & 6.1   & 21.9  & 12.5  & 6.8   & 5.7   & 5.6   & 3.7   & 7.8   &15.1   & 0.9   & 0.0   & 0.3   & 1.4   & 0.6   & 0.8    &0.4\\
9 : Simple negation            & 36.0  & 1.5   & 42.9  & 10.7  & 0.0   & 0.6   & 0.0   & 0.0   & 32.7   &13.0   & 0.3   & 0.0    & 0.0   & 0.0   & 0.0   & 0.0     &0.0\\
10: Indefinite knowledge       & 56.0  & 2.6   & 23.1  & 16.5  & 0.8   & 0.6   & 0.0   & 0.0   & 3.5    &12.9   & 0.0   & 0.0    & 0.2   & 0.0   & 0.0   & 0.0     &0.0\\
 11: Basic coreference          & 38.0  & 3.3   & 4.3   & 0.0   & 11.3  & 0.5   & 0.0   & 0.0   & 0.9   &14.7   & 0.1   & 0.0   & 0.0   & 0.0   & 0.0   & 0.0    &0.0\\
 12: Conjunction                & 26.0  & 0.0   & 3.5   & 0.0   & 0.0   & 0.0   & 0.0   & 0.0   & 0.0   &15.1   & 0.0   & 0.0   & 0.0   & 0.0   & 0.0   & 0.0    &0.0\\
 13: Compound coreference       & 6.0   & 0.5   & 7.8   & 0.0   & 5.3   & 5.5   & 0.0   & 0.3   & 8.9   &13.7   & 0.0   & 0.0   & 0.0   & 0.0   & 0.0   & 0.0    &0.0\\
 14: Time reasoning             & 73.0  & 2.0   & 61.9  & 1.2   & 20.2  & 1.3   & 0.8   & 3.8   & 18.2  &14.5   & 0.1   & 0.0   & 0.0   & 0.2   & 0.0   & 0.0    &0.1\\
 15: Basic deduction            & 79.0  & 1.8   & 47.6  & 0.0   & 39.4  & 0.0   & 0.0   & 0.0   & 0.1   &14.7   & 0.0   & 0.0   & 0.0   & 0.0   & 0.0   & 0.0    &0.0\\
 16: Basic induction            & 77.0  & 51.0  & 54.4  & 0.1   & 50.6  & 54.8  & 53.0  & 53.4  & 53.5  &15.5   & 51.8  & 45.3  & 0.0   & 49.4  & 50.4  & 49.1   &0.0\\
 17: Positional reasoning       & 49.0  & 42.6  & 44.1  & 41.7  & 40.6  & 36.5  & 34.4  & 51.8  & 52.0  &13.0   & 18.6  & 4.2   & 27.8  & 0.9   & 0.0   & 5.8    &4.1\\
 18: Size reasoning             & 48.0  & 9.2   & 9.1   & 9.2   & 8.2   & 8.6   & 7.9   & 8.8   & 47.5  &14.9   & 5.3   & 2.1   & 8.5   & 1.6   & 8.4   & 1.8    &0.7\\
 19: Path finding               & 92.0  & 90.6  & 90.8  & 88.5  & 88.8  & 89.8  & 78.7  & 90.7  & 88.6  &13.6   & 2.3   & 0.0   & 31.0  & 36.1  & 1.0   & 27.9   &0.1\\
 20: Agents motivations         & 9.0   & 0.2   & 2.2   & 0.0   & 0.0   & 0.0   & 0.2   & 0.3   & 5.5   &14.6   & 0.0   & 0.0   & 0.0   & 0.0   & 0.0   & 0.0    &0.0\\
 \hline
 \# Failed                      & 20    & 10    & 16    & 10    & 12    & 8     & 7     &{\bf 5}& 13    &20   & 3     &  1     & 3   & 2     & 2     & 3   &{\bf 0}\\ 
 Average error rates (\%)       & 51.3  & 15.2  & 33.2  & 12.7  & 20.1  & 11.7  &{\bf 9.9}& 11.3  & 20.5  &14.2 & 4.2   & 2.8  & 3.7    & 4.6   & 3.2   & 4.3    &{\bf 0.3}\\
 \hline
\end{tabular}
}
\end{center}
\caption{\small bAbI QA dataset~\citep{babi} error rates (\%) of QRN and previous work: LSTM~\citep{babi}, End-to-end Memory Networks (N2N)~\citep{memN2N}, Dynamic Memory Networks (DMN+)~\citep{DMN+}, Gated End-to-end Memory Networks(GMemN2N)~\citep{perez2016gated}. Results within each task of Differentiable Neural Computer(DNC) were not provided in its paper~\citet{graves2016hybrid}). For QRN, a number in the front (1, 2, 3, 6) indicates the number of layers. A number in the back (200) indicates the dimension of hidden vector, while the default value is 50. `r' indicates that the reset gate is used, and `v' indicates that the gates were vectorized. `*' indicates joint training.}
\label{tab:qa-all}
\end{table}

\begin{table}[ht]
\begin{center}
\resizebox{\columnwidth}{!}{
\begin{tabular}{|l||c|c|c|c|c|c||c|c|c|} 
 \hline
\multirow{3}{*}{Task} & \multicolumn{6}{|c||}{Plain} & \multicolumn{3}{|c|}{With Match} \\
\cline{2-10} & \multicolumn{2}{|c|}{Previous works} & \multicolumn{4}{|c||}{QRN} & \multicolumn{2}{|c|}{Previous works} & \multicolumn{1}{|c|}{QRN} \\
 \cline{2-10}
                                & N2N    & GMemN2N   &1r  &2r   & 2r100 &2rv  & N2N+  & GMemN2N+ & 2r+      \\ 
 \hline
  1: Issuing API calls           & 0.1   &{\bf 0.0}   & 0.02  & {\bf 0.0}   & {\bf 0.0}   & {\bf 0.0}   & {\bf 0.0}   & {\bf 0.0}   & {\bf 0.0}     \\ 
 2: Updating API calls          & {\bf 0.0}   & {\bf 0.0}   & 0.11  & 0.01  & {\bf 0.0}   & {\bf 0.0}   & 1.7   &{\bf 0.0}   &{\bf 0.0}  \\
 3: Displaying options          & 25.1  & 25.1  & {\bf 12.6}  & {\bf 12.6} & {\bf 12.6} &{\bf 12.6}  & 25.1 & 25.1  &{\bf 7.6}   \\
 4: Providing extra information & 40.5  & 42.8  &{\bf 14.3}  & {\bf 14.3} &{\bf 14.3} &{\bf 14.3}  &{\bf 0.0}  &{\bf 0.0}   &{\bf 0.0}    \\
 5: Conducting full dialogs     & 3.9   & 3.7   & 9.4  &{\bf 0.6}  & 0.7  & 0.9  & 6.6   & 2.0   &{\bf 0.0}     \\ 
 \hline
 Average error rates (\%)            & 13.9  & 14.3  & 7.3  & {\bf 5.5}     &{\bf 5.5}       & 5.6      & 6.7   & 5.4   &{\bf 1.5}  \\
 \hline
 \hline
 1 (OOV): Issuing API calls             & 27.7  & 17.6  &{\bf 6.6}  & {\bf 6.6}  &{\bf 6.6}  & {\bf 6.6}  & 3.5   &{\bf 0.0}   &{\bf 0.0}   \\
 2 (OOV): Updating API calls            & 21.1  & 21.1  & 8.5  & 8.4  &{\bf 8.4}   & 8.5  & 5.5   & 5.8   &{\bf 0.0 } \\
 3 (OOV): Displaying options            & 25.6  & 24.7  &{\bf 12.4}  & {\bf 12.4} &{\bf 12.4} & 12.5 & 24.8  & 24.9  &{\bf 7.7}  \\
 4 (OOV): Providing extra information   & 42.4  & 43.0  &{\bf 14.3}  &{\bf 14.3} & 14.4 &14.4  &{\bf 0.0}  &{\bf 0.0}   &{\bf 0.0} \\
 5 (OOV): Conducting full dialogs       & 34.5  & 33.3  & 19.5  & 14.0 & 13.71 &{\bf 13.6} & 22.3  & 20.6  &{\bf 4.0}   \\ 
 \hline
 Average error rates (\%)                    & 30.3  & 27.9  & 12.3 & {\bf 11.1} & {\bf 11.1} &{\bf 11.1}  & 11.2  & 10.3  &{\bf 2.3} \\
 \hline
 \hline
 6: DSTC2 dialog                        & 58.9 & 52.6 & 49.9 & 49.5  &{\bf 48.9}  & 53.8  & 59.0  & 51.3  &{\bf 49.3}  \\
 \hline 
\end{tabular}
}
\end{center}
\caption{ \small bAbI dialog and DSTC2 dialog dataset~\citep{bordes2016learning} average error rates (\%) of QRN and previous work: End-to-end Memory Networks(N2N~\citep{bordes2016learning}) and Gated End-to-end Memory Networks(GMemN2N~\citep{perez2016gated}). For QRN, a number in the front (1, 2, 3, 6) indicates the number of layers and a number in the back (100) indicates the dimension of hidden vector, while the default value is 50. `r' indicates that the reset gate is used, `v' indicates that the gates were vectorized, and `+' indicates that `match' was used. }
\label{tab:dialog-all}
\end{table}

\section{Vector Gate Parallelization}\label{sec:app-b}

For vector gates, we have ${\bf z}_t \in \mathbb{R}^d$ instead of $z_t \in \mathbb{R}$. Therefore the following equation replaces Equation~\ref{eq:exp}:

\begin{equation}\label{eq:v-exp}
    {\bf h}_t 
    = \sum_{i=1}^{t} \exp \left \{
    \begin{pmatrix}
    \sum_{j=i+1}^{t} \log \left(1 - {z_j}^1 \right) \\
    \sum_{j=i+1}^{t} \log \left(1 - {z_j}^2 \right) \\
    \vdots \\
    \sum_{j=i+1}^{t} \log \left(1 - {z_j}^d \right)
    \end{pmatrix}
    \right
    \} \circ
    {\bf z}_i \circ
    \tilde{{\bf h}}_i
\end{equation}

where ${z_j}^k$ is the $k$-th column vector of ${z_j}$.
Let $b_{ij} = \log(1-{z_i}^j) $ for brevity.
Then, we can rewrite Equation~\ref{eq:mat} as following:

\begin{equation}\label{eq:v-mat}
{\begin{pmatrix}
{\bf h}_1^\top \\
{\bf h}_2^\top \\
{\bf h}_3^\top \\
\vdots \\
{\bf h}_T^\top
\end{pmatrix}}^j
 = \left[
 \exp\left\{
    \begin{pmatrix}
        0 & -\infty & -\infty & \ldots & -\infty \\
        b_{2j}  & 0 & -\infty & \ldots & -\infty\\
        b_{2j} + b_{3j}  & b_{3j} & 0 & \ldots & -\infty\\
        \vdots & \vdots & \vdots & \ddots & \vdots\\
        \sum_{k=2}^T b_{kj} & \sum_{k=3}^T b_{kj} & \sum_{k=4}^T b_{kj} &\ldots & 0
    \end{pmatrix}
\right\}
\right]
{\begin{pmatrix}
{\bf z}_1 \circ \tilde{{\bf h}}^\top_1 \\
{\bf z}_2 \circ \tilde{{\bf h}}^\top_2 \\
{\bf z}_3 \circ \tilde{{\bf h}}^\top_3 \\
\vdots \\
{\bf z}_T \circ \tilde{{\bf h}}^\top_T
\end{pmatrix}}^j
\end{equation}

Let ${\bf H} = [{\bf h}_1^\top ; \ldots ; {\bf h}_T^\top]$ 
be a $T$-by-$d$ matrix where the transposes ($\top$) of the column vectors ${\bf h}_t$ are concatenated across row.
We similarly define $\tilde{{\bf H}}$ from $\tilde{{\bf h}}_t$. 
Also, let ${\bf z}=[z_1; \ldots ; z_T]$, and ${\bf B}_d$ be a $T$-by-$T$ matrix where $T$ $[0; b_{2d}; \dots ; b_{Td}]$'s are tiled across the column.

Then Equation~\ref{eq:v-mat} is: 

\begin{equation}\label{eq:v-short}
{\bf H} = \begin{pmatrix}
    \left[ {\bf L} \circ \exp \left( {\bf L}\left[{\bf B}_1 \circ {\bf L}'\right]\right)\right]
        \left[ {\bf Z} \circ \tilde{{\bf H}} \right]^1\\
    \left[ {\bf L} \circ \exp \left( {\bf L}\left[{\bf B}_2 \circ {\bf L}'\right]\right)\right]
        \left[ {\bf Z} \circ \tilde{{\bf H}} \right]^2\\
    \vdots \\
    \left[ {\bf L} \circ \exp \left( {\bf L}\left[{\bf B}_d \circ {\bf L}'\right]\right)\right]
        \left[ {\bf Z} \circ \tilde{{\bf H}} \right]^d
 \end{pmatrix} 
\end{equation}

where ${\bf L}, {\bf L}' \in \mathbb{R}^{T \times T} $ are lower and \emph{strictly} lower triangular matrices of $1$'s are tiled across the column. ${\bf Z} = [{\bf z}_1, \dots , {\bf z}_d] \in \mathbb{R}^{T \times d}$.

\section{Model Details}\label{sec:app-d}

\paragraph{Match.}
While similar in spirit, our `Match' model is slightly different from previous work~\citep{bordes2016learning, perez2016gated}. 
We use answer candidate embedding matrix, and add 2 dimension of 0-1 matrix which expresses whether the answer candidate matches with any word in the paragraph and the question. In other words, the softmax is computed by
$\hat{\bf v} = \mathrm{softmax}\left( {\bf W} [{\bf W}^{(y)};{\bf M}^{(y)}] \hat{\bf y} \right) \in \mathbb{R}^V$, where ${\bf W} \in \mathbb{R}^{d \times d}$ and ${\bf W}^{(y)} \in \mathbb{R}^{V \times (d-2)}$ are trainable weight matrices, and ${\bf M}^{(y)}\in \mathbb{R}^{V \times 2} $ is the 0-1 match matrix.

\section{Visualizations}\label{sec:app-c}

\paragraph{Visualization of Story-based QA.}
Figure~\ref{fig:qa-att-all} shows visualization of models for story-based QA tasks.

In the task 3 (left), the model focuses on the facts that contain 'football' in the first layer, and found out where Mary journeyed to before the bathroom in the second layer. In task 7 (right), the model focuses on the facts that provide information about  the location of Sandra.


\begin{figure}[t]
    \centering
    \resizebox{\columnwidth}{!}{
    \begin{tabular}{|c|ccc|c|}
    \hline
    & \multicolumn{3}{|c|}{Layer 1} & Layer 2\\
    \hline
    Task 3: Three Supporting Facts & $z^1$ & $\overrightarrow{r}^1$ & $\overleftarrow{r}^1$ & $z^2$\\
    \hline
    Mary got the football there. & \cellcolor{red!82} 0.82 & \cellcolor{red!100}1.00 & \cellcolor{red!0}0.0 & \cellcolor{red!6}0.06\\
    John went back to the bedroom. & \cellcolor{red!1}0.01 & \cellcolor{red!00}0.00 & \cellcolor{red!72}0.72 & \cellcolor{red!57}0.57\\
    Mary journeyed to the office. & \cellcolor{red!1}0.01 & \cellcolor{red!04}0.04 & \cellcolor{red!6}0.06 & \cellcolor{red!88}0.88\\
    Mary journeyed to the bathroom. & \cellcolor{red!44}0.44 & \cellcolor{red!00}0.00 & \cellcolor{red!89}0.89 & \cellcolor{red!05}0.05\\
    Mary dropped the football. & \cellcolor{red!62}0.62 & \cellcolor{red!1}0.01 & \cellcolor{red!00}0.00 & \cellcolor{red!03}0.03\\
    \hline
    \multicolumn{5}{|l|}{Where was the football before the bathroom? \hfill office}\\
    \hline
    \end{tabular}
    \begin{tabular}{|c|ccc|c|}
    \hline
    & \multicolumn{3}{|c|}{Layer 1} & Layer 2\\
    \hline
    Task 7: Counting & $z^1$ & $\overrightarrow{r}^1$ & $\overleftarrow{r}^1$ & $z^2$\\
    \hline
    Mary journeyed to the garden. & \cellcolor{red!67} 0.67 & \cellcolor{red!8}0.08 & \cellcolor{red!58}0.58 & \cellcolor{red!12}0.12\\
    Mary journeyed to the office. & \cellcolor{red!91}0.91 & \cellcolor{red!44}0.44 & \cellcolor{red!11}0.11 & \cellcolor{red!21}0.21\\
    Sandra grabbed the apple there. & \cellcolor{red!2}0.02 & \cellcolor{red!34}0.34 & \cellcolor{red!92}0.92 & \cellcolor{red!89}0.89\\
    Sandra discarded the apple. & \cellcolor{red!26}0.26 & \cellcolor{red!61}0.61 & \cellcolor{red!95}0.95 & \cellcolor{red!97}0.97\\
    Daniel went to the bedroom. & \cellcolor{red!70}0.70 & \cellcolor{red!44}0.44 & \cellcolor{red!99}0.99 & \cellcolor{red!3}0.03\\
    \hline
    \multicolumn{5}{|l|}{How many objects is Sandra carrying? \hfill none}\\
    \hline
    \end{tabular}
    }
    \caption{ Visualization of update and reset gates in QRN `2r' model for on several tasks of bAbI QA (Table~\ref{tab:qa-all}).
    We do not put reset gate in the last layer.
    Note that we only show some of recent sentences here, though the stories can have as many as 200+ sentences.
    }
    \label{fig:qa-att-all}
\end{figure}


\paragraph{Visualization of Dialog.}
Figure~\ref{fig:dialog-att-all} shows visualization of models for dialog tasks.

In the first dialog of task 1, the model focuses on the user utterance that mentions the user's desired cuisine and location, and the current query (user's last utterance) informs the system of the number of people, so the system is able to learn that it now needs to ask the user about the desired price range. 
In the second dialog of task 1, the model focuses on the facts that provide information about the requests of the user.
In task 4 (third), the model focuses on what restaurant a user is talking about and the information about the restaurant.


\vspace{-4cm}
\begin{figure*}[t]
    \centering
    
    \resizebox{\columnwidth}{!}{\footnotesize
    \begin{tabular}{|C{11cm}|C{0.6cm}C{0.6cm}C{0.6cm}|C{1cm}|}
    \hline
    & \multicolumn{3}{|c|}{Layer 1} & {Layer 2}\\
    \hline
    Task 1 Issuing API calls& $z^1$ & $\overrightarrow{r}^1$ & $\overleftarrow{r}^1$ & $z^2$\\
    \hline
    Good morning. & \cellcolor{red!12}0.12 & \cellcolor{red!34}0.34 & \cellcolor{red!98}0.98 & \cellcolor{red!20}0.20\\
    Hello what can i help you with today. & \cellcolor{red!97}0.97 & \cellcolor{red!97}0.97 & \cellcolor{red!12}0.12 & \cellcolor{red!12}0.12\\
    Can you book a table in rome with italian cuisine. & \cellcolor{red!0}0.00 & \cellcolor{red!87}0.87 & \cellcolor{red!100}1.00 & \cellcolor{red!100}1.00\\
    I'm on it. & \cellcolor{red!73}0.73 & \cellcolor{red!97}0.97 & \cellcolor{red!38}0.38 & \cellcolor{red!0}0.00\\
    How many people would you in your party. & \cellcolor{red!100}1.00 & \cellcolor{red!100}1.00 & \cellcolor{red!00}0.00 & \cellcolor{red!41}0.41\\
    \hline
    \multicolumn{5}{|l|}{For four people please. \hfill Which price range are you looking for.}\\
    \hline
    \end{tabular}
    }
    
    
    \resizebox{\columnwidth}{!}{\footnotesize
    \begin{tabular}{|C{11cm}|C{0.6cm}C{0.6cm}C{0.6cm}|C{1cm}|}
    \hline
    & \multicolumn{3}{|c|}{Layer 1} & Layer 2\\
    \hline
    Task 1 Issuing API calls & $z^1$ & $\overrightarrow{r}^1$ & $\overleftarrow{r}^1$ & $z^2$\\
    \hline
    Can you make a restaurant reservation for eight in a cheap price range in madrid & \cellcolor{red!0}0.00 & \cellcolor{red!100}1.00 & \cellcolor{red!93}0.93 & \cellcolor{red!100}1.00\\
    I'm on it. & \cellcolor{red!0}0.00 & \cellcolor{red!100}1.00 & \cellcolor{red!74}0.74 & \cellcolor{red!0}0.00\\
    Any preference on a type of cuisine. & \cellcolor{red!0}0.00 & \cellcolor{red!11}0.11 & \cellcolor{red!100}1.00 & \cellcolor{red!1}0.01\\
    I love british food. & \cellcolor{red!0}0.00 & \cellcolor{red!99}0.99 & \cellcolor{red!99}0.99 & \cellcolor{red!57}0.57\\
    Okay let me look into some options for you. & \cellcolor{red!100}1.00 & \cellcolor{red!0}0.00 & \cellcolor{red!00}0.00 & \cellcolor{red!2}0.02\\ 
    \hline
    \multicolumn{5}{|l|}{<SILENCE> \hfill API CALL british madrid eight cheap}\\
    \hline
    \end{tabular}
    }

    \resizebox{\columnwidth}{!}{\footnotesize
    \begin{tabular}{|C{11cm}|C{0.6cm}C{0.6cm}C{0.6cm}|C{1cm}|}
    \hline
    & \multicolumn{3}{|c|}{Layer 1} & Layer 2\\
    \hline
    
    Task 4 Providing extra-information  & $z^1$ & $\overrightarrow{r}^1$ & $\overleftarrow{r}^1$ & $z^2$\\
    
    \hline
    
    resto-paris-expen-spanish-8stars R-phone resto-paris-expen-spanish-8stars-phone & \cellcolor{red!71}0.71 & \cellcolor{red!84}0.84 & \cellcolor{red!99}0.99 & \cellcolor{red!36}0.36\\
    
    resto-paris-expen-spanish-8stars R-address resto-paris-expen-spanish-8stars-address & \cellcolor{red!100}1.00 & \cellcolor{red!99}0.99 & \cellcolor{red!100}1.00 & 
    \cellcolor{red!100}1.00\\
    
    resto-paris-expen-spanish-8stars R-location paris & \cellcolor{red!5}0.05 & \cellcolor{red!1}0.01 & \cellcolor{red!100}1.00 & \cellcolor{red!00}0.00\\
    
    resto-paris-expen-spanish-8stars R-number four & \cellcolor{red!2}0.02 & \cellcolor{red!95}0.95 & \cellcolor{red!97}0.97 & \cellcolor{red!0}0.00\\
    
    resto-paris-expen-spanish-8stars R-price expensive & \cellcolor{red!0}0.00 & \cellcolor{red!5}0.05 & \cellcolor{red!92}0.92 & \cellcolor{red!00}0.00\\
    
    resto-paris-expen-spanish-8stars R-rating 8 & \cellcolor{red!38}0.38 & \cellcolor{red!91}0.91 & \cellcolor{red!100}1.00 & 
    \cellcolor{red!10}0.10\\
    
    What do you think of this option: resto-paris-expen-spanish-8stars & \cellcolor{red!90}0.90 & \cellcolor{red!93}0.93 & \cellcolor{red!99}0.99 & \cellcolor{red!100}1.00\\
    
    Let's do it. & \cellcolor{red!0}0.00 & \cellcolor{red!0}0.00 & \cellcolor{red!100}1.00 & \cellcolor{red!0}0.00\\
    
    Great let me do the reservation. & \cellcolor{red!98}0.98 & \cellcolor{red!99}0.99 & \cellcolor{red!97}0.97 & \cellcolor{red!0}0.00\\
    
    \hline
    
    \multicolumn{5}{|l|}{Do you have its address. \hfill Here it is: resto-paris-expen-spanish-8stars-address} \\

    \hline
    \end{tabular}
    }
    
    \caption{\small Visualization of update and reset gates in QRN `2r' model for on several tasks of bAbI dialog and DSTC2 dialog (Table~\ref{tab:dialog-all}).
    We do not put reset gate in the last layer.
    Note that we only show some of recent sentences here, even the dialog has more sentences.
    }
    \label{fig:dialog-att-all}
\end{figure*}

\end{document}